\documentclass[11pt]{article}

\usepackage[final]{acl}

\usepackage{times}
\usepackage{latexsym}
\usepackage[T1]{fontenc}
\usepackage[utf8]{inputenc}
\usepackage{microtype}
\usepackage{inconsolata}

\usepackage{graphicx}
\usepackage{booktabs}
\usepackage{colortbl}
\usepackage{xcolor}
\usepackage{float}
\usepackage{amsmath}
\usepackage{amssymb}
\usepackage{algorithm}
\usepackage{algpseudocode}

\definecolor{lightgray}{gray}{0.9}

\title{A Model with No Head and Many Thoughts}

\author{
  \textbf{Nikita Koriagin}\textsuperscript{1,*},
  \textbf{Yaroslav Aksenov}\textsuperscript{2,\textdagger},
  \textbf{George Bredis}\textsuperscript{2},
\\
  \textbf{Gleb Gerasimov}\textsuperscript{2},
  \textbf{Nikita Balagansky}\textsuperscript{2},
  \textbf{Daniil Gavrilov}\textsuperscript{2}
\\
  \textsuperscript{1}Yandex, \textsuperscript{2}T-Tech \\
  \small{
    \textsuperscript{*}Work done while at T-Tech.
    \textsuperscript{\textdagger}\textbf{Correspondence:} Yaroslav Aksenov,
    \href{mailto:y.o.aksenov@t-tech.dev}{\texttt{y.o.aksenov@t-tech.dev}}
  }
}

\begin{document}
\maketitle

\begin{abstract}
Large language models decode by projecting hidden states through a large vocabulary head at every step. This operation is computationally costly and forces all reasoning to be expressed in discrete tokens. We introduce Soft Latent Thinking, a method that replaces the LM head during reasoning with a lightweight projector, enabling autoregressive rollout in embedding space where reasoning steps remain continuous rather than tokenized. Experiments on DeepSeek-Qwen-1.5B and LLaMA-3.2-3B show that Soft Latent Thinking consistently improves pass@k across all k while reducing per-step compute during chain-of-thought. Our method achieves the highest pass@32 among all soft-thinking approaches, demonstrating that effective reasoning can be carried out in continuous space without discrete token generation.
\end{abstract}

\begin{figure*}[t]
\centering
\includegraphics[width=\textwidth]{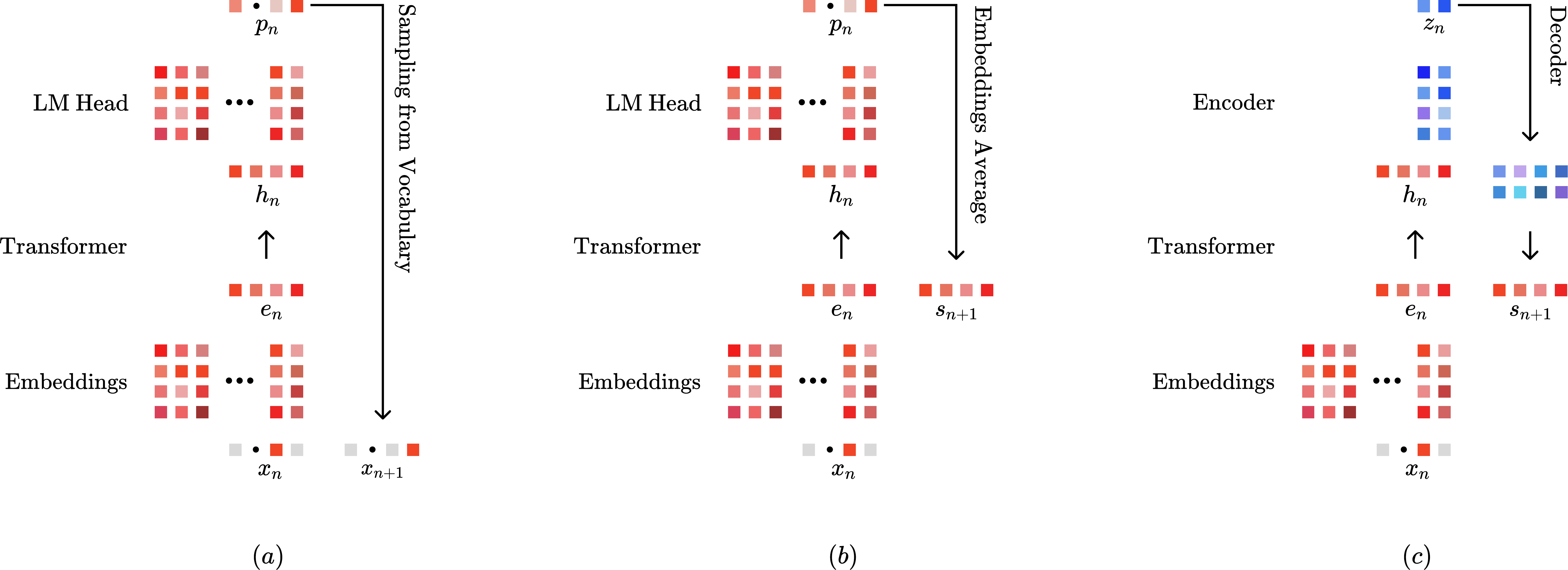}
\caption{Comparison of decoding methods. (a) Standard decoding projects hidden states through the LM head to sample discrete tokens. (b) Soft thinking computes vocabulary probabilities via the LM head, then forms soft embeddings as weighted mixtures over all tokens. (c) Our method replaces the LM head with a lightweight projector (encoder-decoder) producing soft embeddings directly without full vocabulary projection.}

\label{fig:method_comparison}
\end{figure*}
\section{Introduction}

Chain-of-thought (CoT) reasoning improves language model performance on complex problems by allocating extra computation to intermediate steps before producing a final answer.
Soft thinking \cite{softthinking} generalizes CoT by carrying out these intermediate steps in continuous embedding space rather than emitting discrete tokens, enabling richer state transitions and more flexible credit assignment.
Recent methods \cite{hard_truths,softgrpo} further combine soft thinking with reinforcement learning (RL), yielding consistent gains over discrete-token RL on reasoning tasks.

A key limitation of existing soft-thinking approaches is that they still route every reasoning step through the \emph{full} vocabulary head.
Concretely, at each step they form a $V$-way distribution, and use it to produce a weighted mixture of token embeddings.
This design has two undesirable consequences: it (i) ties latent reasoning states to discrete token semantics (since the state must lie in the span of token embeddings), and (ii) keeps the dominant computational cost of the $V$-dimensional projection and normalization, which becomes a bottleneck for long reasoning traces.

We introduce \textbf{Soft Latent Thinking (SLT)}, a soft-thinking variant that replaces the vocabulary projection \emph{only during reasoning} with a compact latent projector.
Instead of producing a distribution over $V$ tokens, SLT maps a hidden state $h_t$ to coefficients over a learned latent basis of size $K \ll V$, and directly synthesizes the next reasoning state in embedding space:
\[
h_t \;\mapsto\; a_t \in \mathbb{R}^{K}, \qquad z_t = B a_t \in \mathbb{R}^{d},
\]
where $B \in \mathbb{R}^{d \times K}$ is the latent basis and $z_t$ is the continuous reasoning state fed to the next step.
For stability, we initialize $B$ from frequent domain token embeddings, but we allow both $B$ and the projector to evolve under RL.
For the final response, we keep standard token decoding unchanged; SLT only alters the internal reasoning operator.
This replacement reduces per-step compute from a $V$-way projection to a $K$-dimensional mapping, and removes the requirement that intermediate reasoning states must be expressible as mixtures of discrete token embeddings.

To make SLT trainable under RL, we define a tractable per-step likelihood for the latent reasoning action via the sampled Gumbel variables used in the continuous selection over the $K$ basis directions.
This provides a well-defined policy for policy-gradient updates despite operating on continuous intermediate states.
In practice, we find that training is reliable with lightweight adaptation: updating only the SLT projector (and basis) or optionally combining it with LoRA~\citep{lora} on a set of backbone weights, without full model retraining.

We evaluate SLT on DeepSeek-R1-Distill-Qwen-1.5B \cite{Deepseek} and LLaMA-3.2-3B-Instruct \cite{Llama} across five mathematical reasoning benchmarks.
Compared to SofT-GRPO, SLT improves average pass@32 while using fewer reasoning tokens and a cheaper per-step reasoning operator.
We also analyze the diversity--accuracy tradeoff induced by the latent operator: individual samples can be slightly less precise than full-vocabulary soft thinking, but the increased diversity across rollouts improves coverage at higher $k$, which primarily drives the pass@32 gains.

\section{Related Work}
Chain-of-thought (CoT) prompting improves language model performance on complex reasoning tasks by encouraging step-by-step intermediate computations before producing final answers \cite{cot,zero_shot}. These works showed that allocating multiple steps of internal reasoning often leads to better outcomes than direct prediction.

Recent work has explored reasoning in continuous embedding space rather than generating discrete tokens at each step. These methods allow models to maintain latent reasoning trajectories without committing to surface forms during intermediate steps. Coconut \cite{coconut} introduces a recurrent hidden-state reasoning loop, where each hidden state is fed back as the next input, while Diffusion-of-Thought \cite{diffusion} refines reasoning through continuous denoising. A distinct class of approaches, known as \emph{soft thinking}, constructs intermediate steps as weighted mixtures of token embeddings \cite{softthinking}. This allows smooth transitions across token semantics but can lead to degenerate deterministic behavior. Wu et al.\ address this by applying Gumbel-Softmax sampling to promote diversity in reasoning paths \cite{softthinking2,gumbel_softmax}.

Recent work applies reinforcement learning with verifiable rewards (RLVR) to enhance CoT reasoning. Group Relative Policy Optimization (GRPO) \citep{grpo} samples groups of trajectories per query and updates the policy to favor higher-reward samples, yielding strong results on reasoning benchmarks. Initial attempts to combine soft thinking with GRPO underperform discrete-token counterparts \citep{hard_truths}, as soft tokens are deterministic given the logits, limiting exploration of alternative reasoning paths. SofT-GRPO \citep{softgrpo} addresses this through Gumbel reparameterization, introducing stochasticity while enabling policy gradients. We build on SofT-GRPO but introduce a dedicated projector that decouples soft token generation from the LM head, reducing per-step compute while improving sample diversity.

\section{Background}

\subsection{Overview}

Standard autoregressive decoding computes at each step:
\begin{equation*}
p(x_t | x_{<t}) = \text{softmax}(W_{\text{head}} \cdot h_t),
\end{equation*}
where $h_t \in \mathbb{R}^d$ is the hidden state and $W_{\text{head}} \in \mathbb{R}^{V \times d}$ projects to the full vocabulary.

\begin{table*}[t]
\centering
\caption{Performance comparison on five mathematical reasoning benchmarks. We evaluate two base models under discrete-token chain-of-thought and soft-thinking reasoning patterns. Metrics @1 denote Mean@32 (average Pass@1 across 32 runs), while @16 and @32 denote Pass@16 and Pass@32 respectively. Best result per metric/dataset is underlined, best average is bolded, second-best average is shaded. All values scaled by 100.}
\label{tab:main_results}
\resizebox{\textwidth}{!}{
\begin{tabular}{l|ccc|ccc|ccc|ccc|ccc|ccc}
\toprule
Dataset & \multicolumn{3}{c|}{AIME2024} & \multicolumn{3}{c|}{AIME2025} & \multicolumn{3}{c|}{AMC23} & \multicolumn{3}{c|}{MATH-500} & \multicolumn{3}{c|}{GSM8K} & \multicolumn{3}{c}{Average} \\
Metrics & @1 & @16 & @32 & @1 & @16 & @32 & @1 & @16 & @32 & @1 & @16 & @32 & @1 & @16 & @32 & @1 & @16 & @32 \\
\midrule
\multicolumn{19}{c}{\textit{DeepSeek-R1-Distill-Qwen-1.5B Base LLM}} \\
\midrule
\multicolumn{19}{c}{Discrete-Token CoT Reasoning Pattern} \\
\midrule
No-Finetune & 30.6 & 70.0 & 73.3 & 23.0 & 46.7 & 53.3 & 70.7 & 92.5 & 95.0 & 84.6 & \underline{97.8} & 97.8 & 81.5 & 95.8 & 96.7 & 58.09 & 80.54 & 83.23 \\
+ GRPO & 31.8 & 66.7 & 76.7 & 25.3 & 46.7 & 46.7 & \underline{77.3} & 95.0 & 95.0 & \underline{87.1} & 97.4 & 97.8 & 84.9 & 95.1 & 95.8 & \cellcolor{lightgray}61.28 & 80.16 & 82.39 \\
\midrule
\multicolumn{19}{c}{Soft-Thinking Reasoning Pattern} \\
\midrule
No-Finetune & 27.3 & 66.7 & 70.0 & 23.8 & 46.7 & 53.3 & 69.9 & 95.0 & 95.0 & 79.4 & 93.2 & 96.6 & 81.0 & 94.6 & 97.1 & 56.28 & 79.23 & 82.41 \\
+ GRPO & 29.2 & 70.0 & 73.3 & 25.4 & 46.7 & 53.3 & 75.8 & 95.0 & 95.0 & 86.3 & 96.8 & \underline{98.2} & 84.9 & 95.6 & 96.4 & 60.31 & 80.81 & 83.26 \\
+ SofT-GRPO & \underline{32.6} & \underline{76.7} & \underline{80.0} & \underline{26.1} & \underline{50.0} & 53.3 & 76.4 & \underline{97.5} & 97.5 & 86.3 & 97.4 & 98.0 & \underline{85.5} & 96.1 & 97.0 & \textbf{61.39} & \textbf{83.54} & \cellcolor{lightgray} 85.18 \\
+ \textbf{Ours} & 28.7 & 74.3 & \underline{80.0} & 20.5 & 48.3 & \underline{56.0} & 72.3 & \underline{97.5} & \underline{100.0} & 83.9 & 96.7 & 97.6 & 81.2 & \underline{96.5} & \underline{97.5} & 57.32 & \cellcolor{lightgray}82.66 & \textbf{86.22} \\
\midrule
\multicolumn{19}{c}{\textit{LLaMA-3.2-3B-Instruct Base LLM}} \\
\midrule
\multicolumn{19}{c}{Discrete-Token CoT Reasoning Pattern} \\
\midrule
No-Finetune & 4.4 & 20.0 & 26.7 & 0.3 & 0.3 & 1.0 & 18.3 & 65.0 & 75.0 & 38.1 & 75.6 & 84.0 & 67.9 & 92.1 & 94.6 & 25.79 & 50.61 & 56.26 \\
+ GRPO & 7.3 & 23.3 & 26.7 & 0.5 & 3.3 & 3.3 & 27.3 & 62.5 & 67.5 & \underline{48.3} & 77.2 & 82.6 & \underline{79.6} & 95.4 & 96.5 & \cellcolor{lightgray}32.60 & 52.35 & 55.32 \\
\midrule
\multicolumn{19}{c}{Soft-Thinking Reasoning Pattern} \\
\midrule
No-Finetune & 3.4 & 16.7 & 16.7 & 0.2 & 6.7 & 6.7 & 17.6 & \underline{70.0} & \underline{77.5} & 36.7 & 76.0 & 81.4 & 66.9 & 91.6 & 94.7 & 24.96 & 52.18 & 55.39 \\
+ GRPO & \underline{8.0} & 20.0 & 23.3 & \underline{0.7} & 3.3 & \underline{10.0} & 27.3 & \underline{70.0} &  75.0 & 47.8 & 76.8 & 81.8 & 79.2 & 94.8 & 96.3 & \cellcolor{lightgray}32.60 & 53.00 & \cellcolor{lightgray}57.28 \\
+ SofT-GRPO & 7.7 & 23.3 & 26.7 & 0.3 & \underline{10.0} & \underline{10.0} & \underline{31.3} & 67.5 & 67.5 & 47.2 & 77.6 & 83.4 & 77.6 & \underline{96.4} & \underline{97.7} & \textbf{32.83} & \textbf{54.96} & 57.06 \\
+ \textbf{Ours} & 7.1 & \underline{28.1} & \underline{36.7} & 0.3 & 5.0 & \underline{10.0} & 19.1 & 64.2 & 75.0 & 39.6 & \underline{78.6} & \underline{84.6} & 66.8 & 95.5 & 97.2 & 26.58 & \cellcolor{lightgray}54.28 & \textbf{60.70}  \\
\bottomrule
\end{tabular}
}
\end{table*}

\subsection{Soft Thinking Methods}

\paragraph{Soft Thinking.} \citet{softthinking} propose generating soft tokens as probability-weighted mixtures of embeddings. At each step:
\begin{equation*}
s_t = \sum_{i=1}^{V} p_{t,i} \cdot e_i,
\end{equation*}
where $e_i$ is the embedding of token $i$, and $p_t = \text{softmax}(W_{\text{head}} \cdot h_t) \in \mathbb{R}^{V}$. The soft token $s_t$ is fed as input to the next step. This enables continuous reasoning without training, but computes a full softmax over $V$ at each step.

\paragraph{Stochastic Soft Thinking.} \citet{softthinking2} identify that vanilla soft thinking suffers from a greedy pitfall---models rely predominantly on the highest-probability token. They introduce Gumbel-Softmax sampling to encourage exploration of diverse reasoning paths. Given log-probabilities $\log p_{t,i} \in \mathbb{R}^V$
over the vocabulary, Gumbel-Softmax sampling computes:
\begin{align}
g_{t,i} &= \log p_{t,i} + \epsilon_{t,i}, \quad \epsilon_{t,i} \sim \text{Gumbel}(0,1), \label{eq:gumbel-perturb} \\
y_{t,i} &= \frac{\exp(g_{t,i} / \tau_g)}{\sum_{j=1}^{V} \exp(g_{t,j} / \tau_g)}, \label{eq:gumbel-softmax}
\end{align}
where $\tau_g$
is a temperature parameter. The soft token is then $s_t = \sum_{i=1}^{V} y_{t,i} \cdot e_i$. 

\paragraph{SofT-GRPO.}
\citet{softgrpo} adapt GRPO-style RL to the \emph{soft-thinking} setting, where each reasoning step outputs a continuous vector $s_t\in\mathbb{R}^d$ instead of a discrete token. Two issues must be solved to make policy gradients workable: (i) plain soft thinking is effectively deterministic given the logits, so there is little exploration of alternative reasoning paths; and (ii) GRPO/PPO~\citep{ppo} needs a per-step log-probability (or importance ratio), but a soft token $s_t$ is not sampled from a categorical distribution. \\
To obtain stochastic but \emph{valid} soft tokens, SofT-GRPO samples a Gumbel--Softmax mixture over the vocabulary and then maps it to an embedding. For a context
$c_t = [\boldsymbol{Q}, (s_1,\ldots,s_{t-1})]$,
rollout uses the behavior policy $\pi_{\theta_{\text{old}}}$ to form token probabilities, injects i.i.d.\ Gumbel noise, and produces a soft token. \\
The central trick for gradient evaluation is that SofT-GRPO does \emph{not} try to assign a probability density directly to $s_t$.
Instead, it assigns likelihood to the \emph{underlying Gumbel variables} (equivalently, to $g_t$).
Given a candidate policy $\pi_\theta$ producing $p_{t,i} = \pi_{\theta}(i \mid c_t)$,
the stored $g_{t,i}$ is distributed as a standard Gumbel shifted by $\log p_{t,i}$, which yields the per-step log-likelihood \\
\begin{align}
\log p_{\theta}(g_t \mid c_t)
&= \sum_{i=1}^{V}\Big[-(g_{t,i}-\log p_{t,i}) \nonumber\\
&\qquad-\exp\!\big(-(g_{t,i}-\log p_{t,i})\big)\Big].
\label{eq:gumbel-likelihood}
\end{align}
Crucially, during the update \emph{$g_{t,i}$ is treated as fixed} because it is part of the sampled trajectory produced in rollout (under $\theta_{\text{old}}$). Equivalently, the underlying noise $\epsilon_{t,i}
= g_{t,i} - \log p_{t,i}^{\text{old}}$
is fixed. Therefore $g_{t,i}$ does \emph{not} depend on $\theta$ inside the gradient computation: the only $\theta$-dependent quantity above is $p_{t,i}=\pi_\theta(i\mid c_t)$.

Finally, SofT-GRPO plugs this likelihood into a GRPO/PPO-style importance ratio for the soft-thinking steps:
\begin{align}
\log r_t
&= \log p_{\theta}(g_t \mid c_t)
- \log p_{\theta_{\text{old}}}(g_t \mid c_t) \nonumber\\
&= \sum_{i=1}^{V}\Big[-(g_{t,i}-\log p_{t,i}) \nonumber\\
&\qquad-\exp\!\big(-(g_{t,i}-\log p_{t,i})\big)\Big] \nonumber\\
&\quad-\sum_{i=1}^{V}\Big[-\epsilon_{t,i}-\exp(-\epsilon_{t,i})\Big].
\label{eq:importance_ratio}
\end{align}
The second sum is constant with respect to $\theta$, so optimizing the GRPO surrogate increases the likelihood of the stored Gumbel variables (and thus the stored soft-thinking trajectory) under $\pi_\theta$ when its final answer receives higher reward.
Answer tokens $\boldsymbol{A}$ are handled exactly as in standard GRPO (categorical log-probabilities and ratios), while only the soft-thinking prefix uses the Gumbel-based ratio above; clipping and a KL penalty to a reference policy are then applied in the usual way to stabilize updates.
\section{Soft Latent Thinking}
\label{sec:method}
Our method, \textit{Soft Latent Thinking}, replaces the expensive vocabulary projection during reasoning with a dedicated \emph{latent projector} that maps the model hidden state directly to a soft embedding.
Unlike SofT-GRPO~\citep{softgrpo}, which forms soft tokens as mixtures over the \emph{full} vocabulary embeddings, our projector learns an independent, lower-cardinality latent space of size $K \ll V$ and decodes it back to $\mathbb{R}^d$.
This decoupling allows soft tokens to carry information that is not constrained to discrete token semantics, while reducing per-step compute by operating over $K\approx 12\text{k}$--$24\text{k}$ instead of $V\approx 150\text{k}$.

At soft-thinking step $t$, let $h_t\in\mathbb{R}^d$ denote the last-layer hidden state (before the LM head) produced by the base model given the current context.
The projector produces a soft embedding $s_t\in\mathbb{R}^d$ via a compressed softmax and Gumbel--Softmax sampling:

\paragraph{Encoder.}
The encoder is a single linear map that compresses the hidden state into $K$ logits:
\begin{equation*}
z_t
= W_{\text{enc}}\, h_t.
\end{equation*}

\paragraph{Sampling.}
We first compute probabilities from the encoder logits:
\begin{align}
p_{t,i} = \frac{\exp(z_{t,i})}{\sum_{j=1}^{K}\exp(z_{t,j})}.
\end{align}
We then apply Gumbel-Softmax (Eqs.~\ref{eq:gumbel-perturb}--\ref{eq:gumbel-softmax}) over $K$ categories, yielding mixture weights $y_t \in \mathbb{R}^K$.

\paragraph{Decoder.}
The decoder maps the latent mixture $y_t$ back to the model embedding space:
\begin{align}
s_t
&= W_{\text{dec}}^{\top} y_t.
\end{align}

\begin{algorithm}[t]
\small
\caption{Soft Latent Thinking training step.}
\label{alg:slt}
\begin{algorithmic}[1]
\Require prompt $Q$, policies $\theta_{\rm old},\theta$, projectors $(W_{\rm enc}^{old},W_{\rm dec}^{old})$ and $(W_{\rm enc},W_{\rm dec})$, temperature $\tau_g$
\Statex \textbf{Rollout}
\State $c_1\gets Q$, \quad $\mathcal{B}\gets\emptyset$
\For{$t=1,\ldots,T_{\max}$}
  \State $h_t\gets f_{\theta_{\rm old}}(c_t)$
  \State $p_t\gets\mathrm{softmax}(W_{\rm enc}^{old}h_t)$
  \State $\epsilon_t\sim\mathrm{Gumbel}(0,1)^K$, \quad $g_t\gets\log p_t+\epsilon_t$
  \State $y_t\gets\mathrm{softmax}(g_t/\tau_g)$
  \State $s_t\gets (W_{\rm dec}^{old})^\top y_t$
  \State $\mathcal{B}\gets\mathcal{B}\cup\{(c_t,g_t,\epsilon_t,s_t)\}$, \quad $c_{t+1}\gets[c_t;s_t]$
  \If{$\mathrm{stop}(s_t)$} \Comment{aligned with \texttt{</think>} or \texttt{\textbackslash boxed}}
    \State \textbf{break}
  \EndIf
\EndFor
\State Decode answer with LM head; receive reward $R$
\Statex \textbf{Update}
\For{$(c_t,g_t,\epsilon_t,s_t)\in\mathcal{B}$}
  \State $h_t^\theta\gets f_\theta(c_t)$
  \State $p_t^\theta\gets\mathrm{softmax}(W_{\rm enc}h_t^\theta)$
  \State $\log r_t\gets\log p_\theta(g_t\mid c_t)-\log p_{\theta_{\rm old}}(g_t\mid c_t)$ \Comment{Eq.~\ref{eq:gumbel-likelihood}}
\EndFor
\State Optimize clipped GRPO objective using $\{\log r_t\}$, answer-token ratios, and $R$
\end{algorithmic}
\end{algorithm}

\subsection{Initialization}
Let $\mathcal{K}=\{k_1,\ldots,k_K\}$ be the indices of the $K$ most frequent tokens in the target domain (or another chosen token subset).
We initialize the projector from the pretrained LM head $W_{\text{head}}\in\mathbb{R}^{V\times d}$ and the input embedding table $E\in\mathbb{R}^{V\times d}$ by copying the corresponding rows:
\begin{align}
\label{init}
W_{\text{enc}}[i,:]
&\leftarrow W_{\text{head}}[k_i,:],
\\
W_{\text{dec}}[i,:]
&\leftarrow E[k_i,:],
\qquad
i\in\{1,\ldots,K\}.
\end{align}
After initialization, $W_{\text{enc}}$ and $W_{\text{dec}}$ are trained and are not tied to $W_{\text{head}}$ or $E$.

\begin{figure}[t]
\centering
\includegraphics[width=\columnwidth]{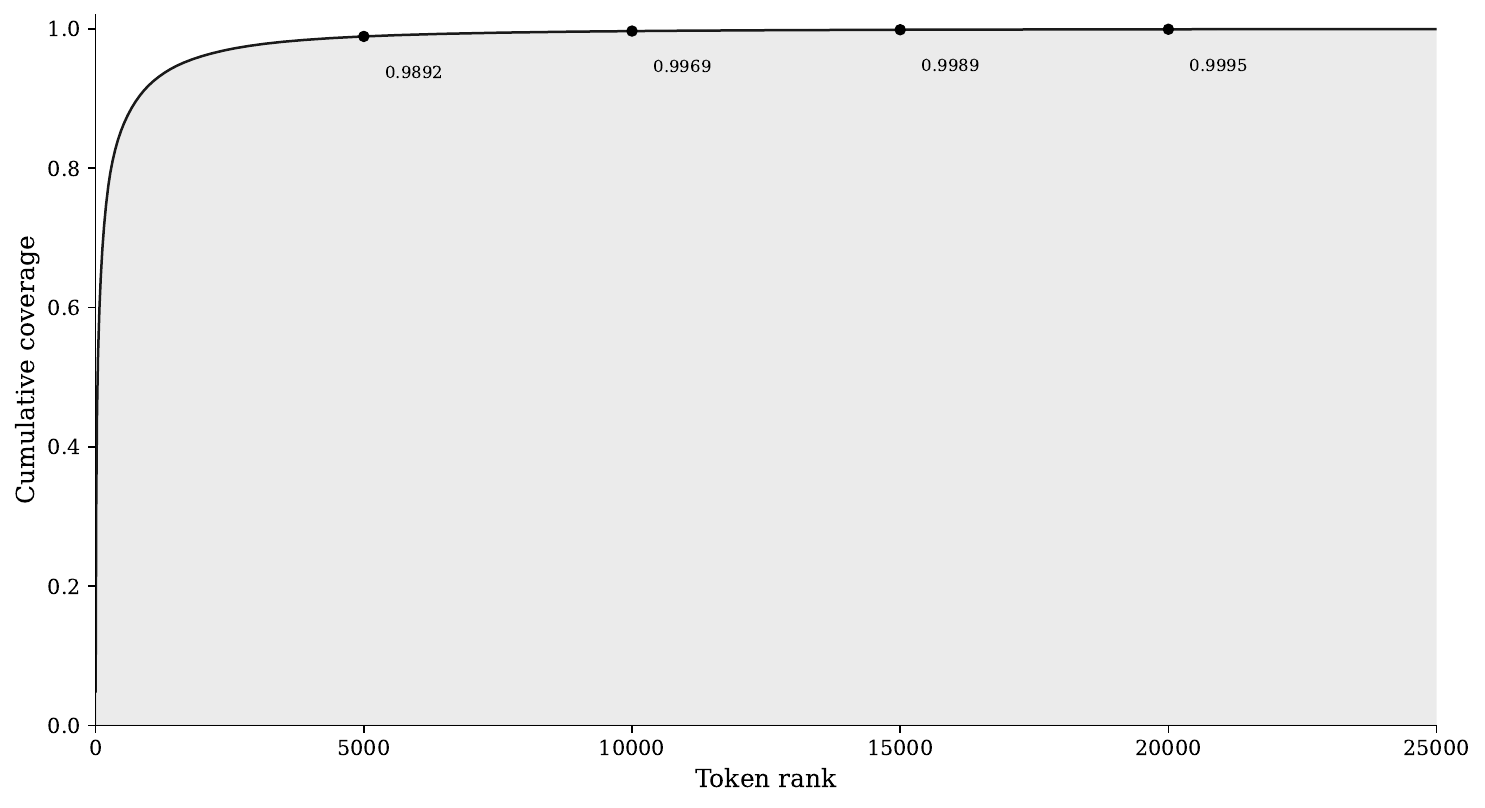}
\caption{Cumulative distribution of token frequencies on mathematical reasoning data. A small subset of tokens covers most occurrences: the top 5000 tokens account for almost 99\% of the distribution.}
\label{fig:token_cdf}
\end{figure}

The token-frequency distribution used for initialization is shown in Figure~\ref{fig:token_cdf}; it is computed on mathematical reasoning data (the math subset of OpenThoughts-114k \cite{open_thoughts}, containing DeepSeek-generated \cite{Deepseek} reasoning traces). The distribution is highly concentrated, but although the top 5k tokens cover over 99\% of occurrences, our ablations (Table~\ref{tab:ablation_scale}) show that $K \approx 12$k yields optimal performance, suggesting the model benefits from access to rarer tokens during reasoning.

\begin{table*}[t]
\centering
\caption{Token efficiency on mathematical reasoning benchmarks . \#Token represents total tokens generated across all queries, while \#Token$_c$ represents tokens generated for correctly solved queries only.}

\label{tab:tokens}
\resizebox{\textwidth}{!}{
\begin{tabular}{l|cc|cc|cc|cc|cc|cc}
\toprule
Dataset & \multicolumn{2}{c|}{AIME2024} & \multicolumn{2}{c|}{AIME2025} & \multicolumn{2}{c|}{AMC23} & \multicolumn{2}{c|}{MATH-500} & \multicolumn{2}{c|}{GSM8K} & \multicolumn{2}{c}{Average} \\
Metrics & \#Token & \#Token\_c & \#Token & \#Token\_c & \#Token & \#Token\_c & \#Token & \#Token\_c & \#Token & \#Token\_c & \#Token & \#Token\_c \\
\midrule
\multicolumn{13}{c}{\textit{DeepSeek-R1-Distill-Qwen-1.5B Base LLM}} \\
\midrule
\multicolumn{13}{c}{Discrete-Token CoT Reasoning Pattern} \\
\midrule
No-Finetune & 16241.6 & 14997.8 & 16416.3 & 13448.8 & 10052.2 & 9394.2 & 5616.6 & 5368.1 & 1839.3 & 1772.5 & 10033.2 & 8996.3 \\
+ GRPO & 8927.6 & 8535.6 & 8039.4 & 6414.2 & 5001.3 & 4672.8 & 3106.1 & 2958.5 & 1417.1 & 1370.6 & 5298.3 & 4790.3 \\
\midrule
\multicolumn{13}{c}{Soft-Thinking Reasoning Pattern} \\
\midrule
No-Finetune & 17857.8 & 16191.3 & 17569.4 & 14582.4 & 11269.9 & 10482.0 & 4015.9 & 3888.3 & 1699.3 & 1649.9 & 10482.5 & 9358.8 \\
+ GRPO & 9383.2 & 8934.5 & 8007.2 & 7325.8 & 5203.7 & 4894.4 & 3233.6 & 3131.7 & 1385.5 & 1349.9 & 5442.7 & 5127.2 \\
+ SofT-GRPO & 11039.6 & 10756.1 & 10519.6 & 7831.3 & 5900.2 & 5630.4 & 3549.5 & 3399.2 & 1577.6 & 1542.5 & 6517.3 & 5831.9 \\
+ \textbf{Ours} & 10280.8 & 7006.2 & 9805.2 & 7630.9 & 5252.6 & 4423.4 &3379.5 &2815.0 & 1646.9 & 1342.6 &6073.0 & 4643.6 \\
\bottomrule
\end{tabular}
}
\end{table*}
\subsection{Training}
We train with the SofT-GRPO objective~\citep{softgrpo}.
During rollout, we sample soft-thinking trajectories using the  policy parameters $\theta_{\text{old}}$.
At each soft-thinking step $t$, we compute  encoder logits  ${z}^{old}_t=W_{enc}(\theta_{old})h_t$, apply Gumbel-Softmax sampling, and store the perturbed log probs $g_t$ together with the decoder produced soft embedding $s_t$ (which is used as the next-step input).
During the policy update, we replay the stored soft embeddings $(s_1,\ldots,s_{t-1})$ to form the same context and recompute the projector probabilities under the \emph{current} parameters $\theta$:
\begin{align*}
z_t
&= W_{\text{enc}}(\theta)\, h_t,
\\
p_{t,i}
&= \frac{\exp(z_{t,i})}
        {\sum_{j=1}^{K}\exp(z_{t,j})}.
\end{align*}
Following SofT-GRPO, we evaluate the likelihood of the \emph{stored} Gumbel-perturbed log probs $g_t=(g_{t,1},\ldots,g_{t,K})$ under the current policy and evaluate the Gumbel likelihood (Eq.~\ref{eq:gumbel-likelihood} with $V$ replaced by $K$).
In this computation, $g_{t,i}$ is treated as \emph{fixed} because it is part of the sampled trajectory produced during rollout (under $\theta_{\text{old}}$); only $p_{t,i}$ depends on $\theta$.
The corresponding per-step importance ratio used by the GRPO surrogate is
\begin{align}
\log r_t
&=
\log p_{\theta}(g_t \mid c_t)
-
\log p_{\theta_{\text{old}}}(g_t \mid c_t),
\end{align}
where the second term is constant w.r.t.\ $\theta$ for a fixed rollout sample.
For the final answer tokens, we use the standard categorical log-probability ratios from the LM head, as in discrete-token GRPO.

\subsection{Inference}
Generation begins with a \texttt{<think>} token when available.
At each reasoning step, the projector produces a soft embedding $s_t$, which is fed to the next model step.
Reasoning terminates when the soft embedding is sufficiently aligned with the \texttt{</think>} token embedding:
\begin{align}
\cos\!\big(s_t,\, e_{\texttt{</think>}}\big)
&> \delta .
\end{align}
The model then switches to standard autoregressive decoding to generate the final answer.
For models without explicit thinking boundary tokens (e.g., some LLaMA-style checkpoints), we use an alternative stopping criterion. Reasoning terminates when the soft embedding is sufficiently close to the \texttt{\textbackslash boxed} token embedding:
\begin{equation}
\cos(s_t, e_{\texttt{\textbackslash boxed}}) > \gamma.
\end{equation}
\paragraph{Stopping-threshold sensitivity.}
The stopping rule uses a cosine-similarity threshold to decide when the latent reasoning phase should hand control back to ordinary token decoding. In preliminary sensitivity checks, we observed that the cosine similarity to the boundary embedding stays low during intermediate reasoning, typically below $0.2$, and rises sharply near the point where the model begins to produce the final answer. This makes the stopping rule relatively insensitive to moderate changes in $\delta$ or $\gamma$, although a full sweep over thresholds remains an important robustness check.

\section{Experiments}
\label{sec:experiments}
We implement Soft Latent Thinking on two base models: DeepSeek-R1-Distill-Qwen-1.5B \citep{Deepseek} and LLaMA-3.2-3B-Instruct \citep{Llama}. 

\paragraph{Projector.} The projector consists of an encoder $W_{\text{enc}} \in \mathbb{R}^{K \times d}$ and decoder $W_{\text{dec}} \in \mathbb{R}^{K \times d}$. We set $K = 8d$, giving $K = 12288$ for Qwen ($d = 1536$) and $K = 24576$ for LLaMA ($d = 3072$). 
\paragraph{Training.} We train on the DeepScaleR dataset \citep{deepscaler} using Soft-GRPO with outcome-based rewards. For both models we freeze the backbone and train LoRA adapters (rank 64) on all attention and MLP modules jointly with the projector.

\paragraph{Baselines.} We compare against: (1) the base model without fine-tuning, (2) the base model trained with standard GRPO, (3) Soft-Thinking without fine-tuning \citep{softthinking}, (4) Soft-Thinking with GRPO, and (5) SofT-GRPO \citep{softgrpo}.

\subsection{Main Results}

The main results of Soft Latent Thinking are shown in Table~\ref{tab:main_results}. We evaluate on five mathematical reasoning benchmarks: AIME2024, AIME2025~\cite{aime}, AMC23~\cite{amc}, MATH-500~\cite{math}, and GSM8K~\cite{gsm8k}.

Soft Latent Thinking primarily improves the multi-sample accuracy--efficiency tradeoff rather than uniformly dominating at every sampling budget. On DeepSeek-R1-Distill-Qwen-1.5B, we achieve 86.22 average pass@32 compared to 83.23 for the base model and 85.18 for SofT-GRPO. On LLaMA-3.2-3B-Instruct, we achieve 60.70 average pass@32 compared to 56.26 for the base model and 57.06 for SofT-GRPO.

The gains are most pronounced at higher $k$, suggesting that decoupling soft token generation from the LM head---combined with Gumbel-Softmax sampling---encourages more diverse reasoning paths across rollouts. This also clarifies the deployment regime: SLT is most useful when multiple rollouts are affordable or already required, such as RL training and pass@$k$ inference.

\subsection{Preliminary Larger-Model Check}
\label{sec:qwen35_9b}
To probe whether the behavior transfers beyond the 1.5B--3B setting, we additionally ran a single untuned experiment on Qwen3.5-9B~\citep{qwen35}. This check is not intended as a full-scale evaluation, because we did not perform scale-specific hyperparameter tuning or a complete baseline sweep. Nevertheless, as Table~\ref{tab:qwen35_9b} shows, it reproduces the qualitative pattern observed in smaller models: the projector model is weaker at low $k$, but reaches the same pass@32 while using substantially fewer reasoning tokens.

\begin{table}[t]
\centering
\small
\caption{Preliminary AIME2024 larger-model sanity check on Qwen3.5-9B. Results are from one untuned configuration and should be interpreted as evidence of transfer stability rather than as a definitive scaling study.}
\label{tab:qwen35_9b}
\resizebox{\columnwidth}{!}{%
\begin{tabular}{lccccc}
\toprule
Model & Avg. tokens & Correct tokens & Pass@1 & Pass@16 & Pass@32 \\
\midrule
Qwen3.5-9B + projector & 9288.1 & 7378.2 & 68.4 & 92.8 & 93.3 \\
Base Qwen3.5-9B & 23292.1 & 21778.6 & 76.4 & 93.2 & 93.3 \\
\bottomrule
\end{tabular}%
}
\end{table}

\subsection{Out-of-Domain Evaluation}

\begin{figure*}[t]
  \centering
  \includegraphics[width=0.48\linewidth]{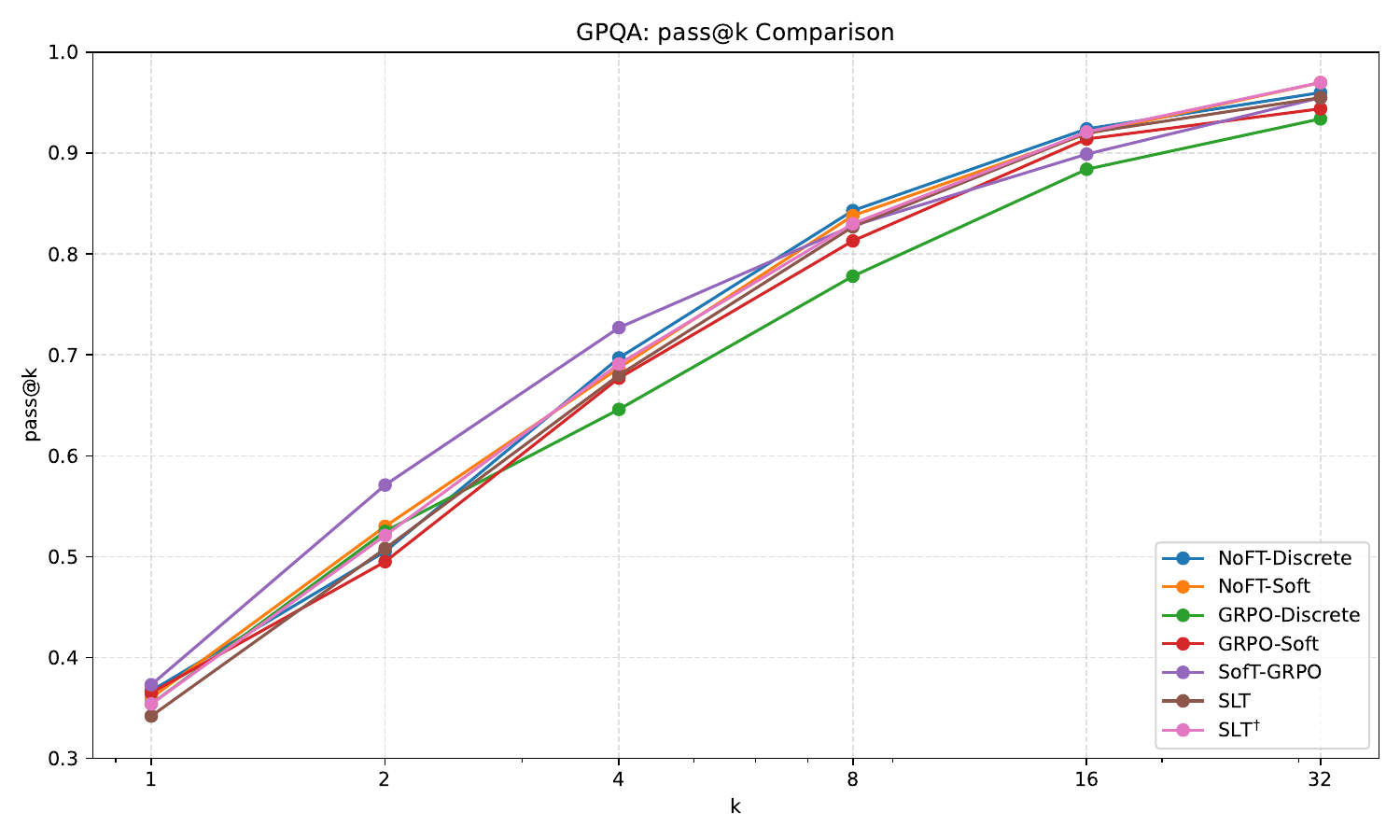}
  \hfill
  \includegraphics[width=0.48\linewidth]{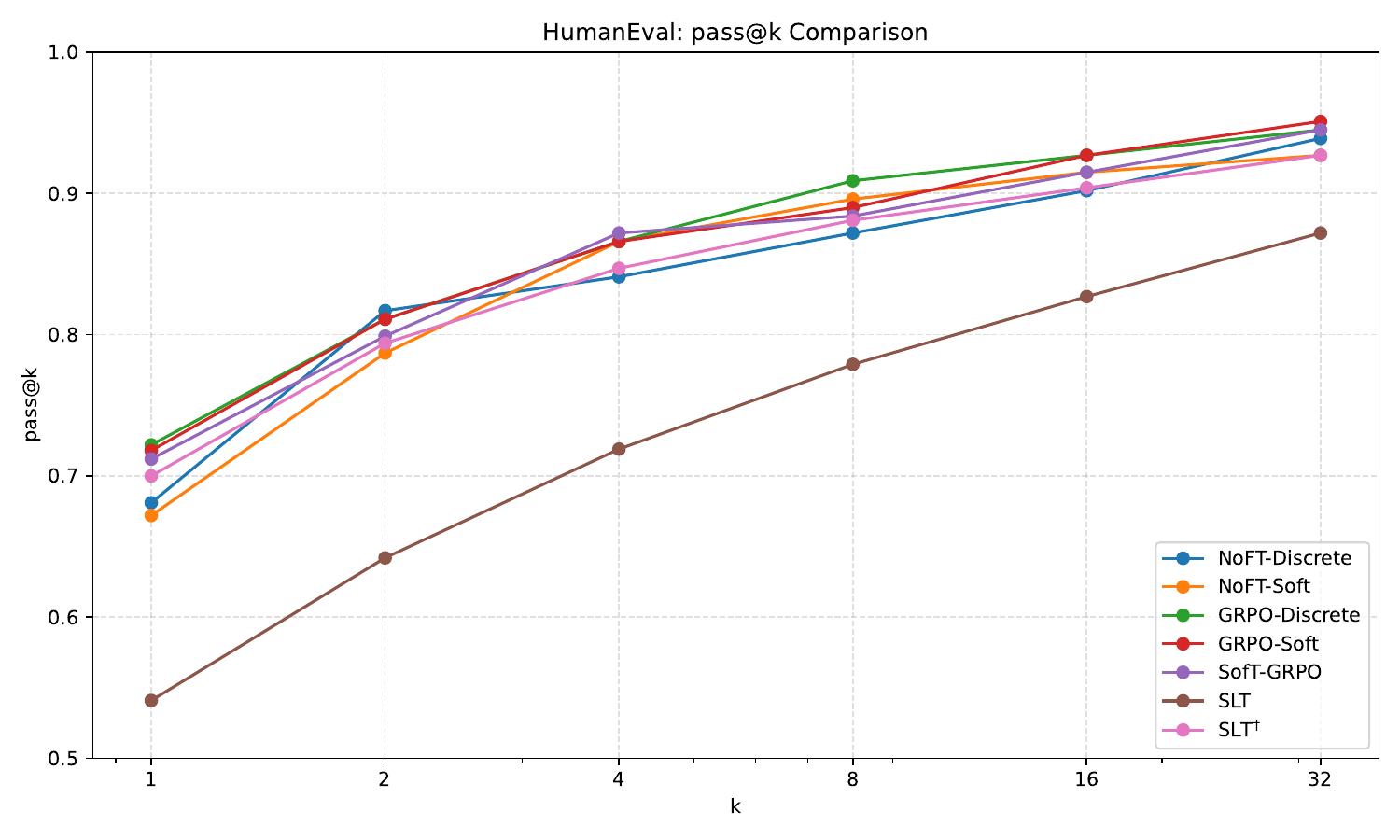}
  \caption{Out-of-domain evaluation on GPQA Diamond (science reasoning) and HumanEval (code generation) using \mbox{DeepSeek-R1-Distill-Qwen-1.5B} trained on mathematical reasoning tasks. SLT$^{\dagger}$ indicates the projector is disabled at inference time, reverting to standard soft thinking with full-vocabulary decoding.}
  \label{fig:passatk-comparison}
\end{figure*}

We evaluate out-of-domain behavior on GPQA Diamond (science) \cite{gpqa} and HumanEval (code) \cite{humaneval}. The corresponding pass@$k$ curves are shown in Figure~\ref{fig:passatk-comparison}, and the full numerical results are reported in Table~\ref{tab:ood}. We evaluate three settings: base model, LoRA with projector enabled, and LoRA with standard soft thinking (full vocabulary, no projector).

On GPQA, the projector transfers well, achieving accuracy comparable to baselines. We attribute this to vocabulary overlap between mathematical and scientific reasoning (shared use of numbers, symbols, and formal notation).

On HumanEval, the math-initialized projector degrades performance, as code requires different vocabulary (keywords, syntax, identifiers). However, disabling the projector and using standard soft thinking with the jointly trained LoRA achieves performance comparable to the strongest baselines. This demonstrates that LoRA adapters trained jointly with the projector remain effective for out-of-domain tasks when the projector is bypassed. Mechanistically, the degradation is expected: code generation relies on Python keywords, identifiers, syntax tokens, and indentation markers, whereas the projector is initialized from tokens frequent in mathematical reasoning.

Crucially, when the projector is disabled, performance matches or exceeds SofT-GRPO on out-of-domain tasks (97.0 vs 95.5 on GPQA, 92.7 vs 94.5 on HumanEval). This enables a practical deployment strategy: use the projector for in-domain tasks to gain efficiency and pass@k improvements, and disable it for out-of-domain tasks with no performance penalty relative to the strongest baseline.

The full numerical results are reported in Table~\ref{tab:ood}.

\subsection{Token Efficiency}

Table~\ref{tab:tokens} presents token usage across benchmarks. Soft Latent Thinking uses fewer tokens on average compared to SofT-GRPO, while achieving higher pass@$k$ at higher $k$. This suggests that reasoning in continuous embedding space allows for more compact reasoning chains without sacrificing accuracy. The reduced token count, combined with the lower per-step FLOPs from bypassing the full vocabulary projection, results in overall computational savings during chain-of-thought generation.

\subsection{Computational Efficiency}

The primary advantage of Soft Latent Thinking is reduced compute during chain-of-thought generation. At each reasoning step, the standard approach computes:
\begin{equation*}
\text{LM head FLOPs} = 2 \times d \times V
\end{equation*}
Our projector instead computes:
\begin{equation*}
\text{Projector FLOPs} = 2 \times d \times K + 2 \times K \times d = 4dK
\end{equation*}
For $d = 1536$, $V = 150$k, and $K = 16$k, this yields a reduction of $\frac{2dV}{4dK} = \frac{V}{2K} \approx 5 \times$ on the vocabulary projection step.

We also measured prototype serving throughput in a mini-SGLang~\citep{sglang} implementation. To be conservative, Table~\ref{tab:serving_speed} reports projector scale 8, the largest and least favorable projector among the tested settings. Even in this setting, throughput remains above the vanilla baseline across the tested models and batch sizes. We separate these end-to-end serving numbers from graph-level decode speed: the latter shows an approximately $1.05\times$ decode speedup across tested models and is the cleaner measurement of the algorithmic gain from replacing the LM-head reasoning step.

\begin{table}[H]
\centering
\small
\caption{Prototype serving throughput with projector scale 8. TPS denotes generated tokens per second.}
\label{tab:serving_speed}
\resizebox{\columnwidth}{!}{%
\begin{tabular}{lrrrr}
\toprule
Model & Batch & Vanilla TPS & Ours TPS & Speedup \\
\midrule
Llama-3.2-1B & 1 & 675.4 & 723.3 & 1.071$\times$ \\
Llama-3.2-1B & 4 & 2637.1 & 2823.1 & 1.071$\times$ \\
Llama-3.2-1B & 8 & 5038.2 & 5349.2 & 1.062$\times$ \\
Llama-3.2-3B & 1 & 299.3 & 314.7 & 1.051$\times$ \\
Llama-3.2-3B & 4 & 1176.7 & 1236.2 & 1.051$\times$ \\
Llama-3.2-3B & 8 & 2317.8 & 2436.6 & 1.051$\times$ \\
Llama-3.1-8B & 1 & 160.2 & 165.3 & 1.032$\times$ \\
Llama-3.1-8B & 4 & 519.9 & 652.9 & 1.256$\times$ \\
Llama-3.1-8B & 8 & 1113.7 & 1266.1 & 1.137$\times$ \\
\bottomrule
\end{tabular}%
}
\end{table}

\subsection{Ablations}
All ablation studies are conducted on AIME2024 using DeepSeek-R1-Distill-Qwen-1.5B unless otherwise specified. 
\paragraph{Effect of projector training.} We ablate the projector by replacing trained weights with initial ones (Eq.~\ref{init}) while keeping LoRA fixed. As shown in Table~\ref{tab:ablation_projector}, this degrades performance across all pass@$k$ and increases response length, indicating the projector learns more efficient reasoning paths despite minimal weight change.

We also compare our projector against standard soft thinking using the same LoRA weights (Table~\ref{tab:ablation_projector}). Full vocabulary soft thinking achieves higher pass@1 but lower pass@16, with similar pass@32. This suggests the compressed vocabulary introduces beneficial stochasticity: while individual samples may be less precise, diversity across samples improves, yielding better coverage at higher $k$.

\begin{table}[H]
\centering
\small
\caption{Projector variant ablation with fixed LoRA weights. Full ST uses standard soft thinking over full vocabulary $V$.}
\label{tab:ablation_projector}
\resizebox{\columnwidth}{!}{%
\begin{tabular}{l|ccc|c}
\toprule
Setting & @1 & @16 & @32 & \#Token \\
\midrule
Ours (trained projector) & 28.7 & 74.3 & 80.0 & 10280 \\
Init projector& 28.3 & 70.1 & 76.7 & 11383  \\
Full ST  & 29.5 & 72.3 & 80.0 & 11242  \\
\bottomrule
\end{tabular}
}
\end{table}

\paragraph{Vocabulary compression at inference.} 
We evaluate vocabulary compression without training by using an initialized projector (pruned LM head and embedding table to $K$ tokens) at inference time. Table~\ref{tab:compression} compares this against the full vocabulary baseline.

Without fine-tuning, the pruned projector degrades performance compared to full vocabulary soft thinking, indicating that vocabulary compression alone hurts. When applied to the SofT-GRPO checkpoint, the pruned projector shows the same degradation. This demonstrates that vocabulary compression at inference alone is insufficient---the model must be trained with the compressed vocabulary to effectively reason through the bottleneck.

\begin{table}[H]
\centering
\small
\caption{Vocabulary compression at inference without training.}
\label{tab:compression}
\resizebox{\columnwidth}{!}{%
\begin{tabular}{l|ccc|c}
\toprule
Setting & @1 & @16 & @32& \#Token \\
\midrule
\multicolumn{5}{c}{\textit{No fine-tuning}} \\
\midrule
Full ST  & 27.3 & 66.7 & 70.0 &  17858 \\
Init projector & 11.3 & 56.3 & 63.3 &  6370 \\
\midrule
\multicolumn{5}{c}{\textit{SofT-GRPO checkpoint}} \\
\midrule
Full ST  & 32.6 & 76.7 & 80.0 & 11039 \\
Init projector & 15.8 & 60.6 & 70.0 &  6408 \\
\bottomrule
\end{tabular}
}
\end{table}

\paragraph{Projector size.} 
Table~\ref{tab:ablation_scale} ablates the projector size $K$. Small $K \approx 1.5$k underperforms due to limited capacity. $K \approx 6$k achieves the best pass@1, while $K \approx 12$k achieves the best pass@16 and pass@32. This suggests a trade-off: smaller projectors produce more focused selections benefiting single attempts, while larger projectors enable greater diversity for multi-sample evaluation.

\begin{table}[H]
\centering
\small
\caption{Ablation on projector size $K$.}
\label{tab:ablation_scale}
\resizebox{\columnwidth}{!}{%
\begin{tabular}{l|ccc|c}
\toprule
$K$ & @1 & @16 & @32 & \#Token \\
\midrule
1*1536 & 20.4 & 58.2 & 66.7 & 18694\\
4*1536 & 29.7 & 62.0 & 66.7 & 14881\\
8*1536 & 28.7 & 74.3 & 80.0 & 10280\\
\bottomrule
\end{tabular}
}
\end{table}

\paragraph{Projector and backbone training.} 
Table~\ref{tab:ablation_training} ablates the training configuration. Training only the projector with a frozen backbone yields weaker performance than the base model, as the backbone cannot adapt to process soft embeddings. Joint training of projector and LoRA achieves the best results, allowing both components to co-adapt. We do not evaluate freezing the projector while training only LoRA, as this reduces to SofT-GRPO constrained to $K$ tokens---strictly less expressive than full vocabulary SofT-GRPO.

\begin{table}[H]
\centering
\small
\caption{Projector and backbone training ablation.}
\label{tab:ablation_training}
\resizebox{\columnwidth}{!}{%
\begin{tabular}{ll|ccc|c}
\toprule
Projector & Backbone & @1 & @16 & @32 & \#Token \\
\midrule
Trained & LoRA & 28.7 & 74.3 & 80.0 & 10280 \\
Trained & Frozen & 26.0 & 65.4 & 70.0 & 12256\\
Init & Frozen &  11.3 & 56.3 & 63.3 & 6370 \\
\bottomrule
\end{tabular}
}
\end{table}

\paragraph{Temperature.}
We explored the Gumbel-Softmax temperature $\tau_g$ during training and inference. For training, $\tau_g=0.1$ worked best for LLaMA-3.2-3B-Instruct while $\tau_g=0.5$ worked best for DeepSeek-R1-Distill-Qwen-1.5B, suggesting that the optimal training temperature may depend on the base model. For inference, both models performed best with $\tau_g=0.5$, indicating that moderate stochasticity during generation encourages exploration of diverse reasoning paths.

\section{Discussion}

\paragraph{Training efficiency.} Replacing the full vocabulary projection ($V \approx 150$k) with a compressed projector ($K \approx 12$--$24$k) reduces this operation by $\sim$5--10$\times$ per reasoning step. End-to-end gains are smaller because attention and MLP layers remain, but the savings compound over long traces and many GRPO rollouts.
\paragraph{Decoupled embeddings.} Standard soft thinking uses the embedding table both for latent reasoning and discrete output, which may constrain expressiveness. Our projector decouples these roles: its decoder learns embeddings optimized for latent reasoning rather than token identity.
\paragraph{LoRA as decoding adapter.} LoRA adapts the frozen backbone to process projector-generated soft embeddings instead of discrete token embeddings. This helps explain why out-of-domain knowledge is preserved when the projector is disabled.

\begin{table}[tbp]
\centering
\small
\caption{Out-of-domain evaluation on GPQA Diamond and HumanEval. $\dagger$ indicates projector disabled, reverting to full-vocabulary soft thinking.}
\label{tab:ood}
\resizebox{\columnwidth}{!}{%
\begin{tabular}{l|ccc|ccc}
\toprule
Dataset & \multicolumn{3}{c|}{GPQA Diamond} & \multicolumn{3}{c}{HumanEval} \\
Metrics & @1 & @8 & @32 & @1 & @8 & @32 \\
\midrule
\multicolumn{7}{c}{\textit{DeepSeek-R1-Distill-Qwen-1.5B}} \\
\midrule
\multicolumn{7}{c}{Discrete-Token CoT Reasoning} \\
\midrule
No-Finetune & 36.7 & \underline{84.3} & 96.0 & 68.1 & 87.2 & 93.9 \\
+ GRPO & 35.4 & 77.8 & 93.4 & \underline{72.2} & \underline{90.9} & 94.5 \\
\midrule
\multicolumn{7}{c}{Soft-Thinking Reasoning} \\
\midrule
No-Finetune & 36.0 & 83.8 & \underline{97.0} & 67.2 & 89.6 & 92.7 \\
+ GRPO & 36.5 & 81.3 & 94.4 & 71.8 & 89.0 & \underline{95.1} \\
+ SofT-GRPO & \underline{37.3} & 82.8 & 95.5 & 71.2 & 88.4 & 94.5 \\
+ \textbf{Ours} & 34.2 & 82.7 & 95.5 & 54.0 & 77.9& 87.2 \\
+ \textbf{Ours$^\dagger$} & 35.4 & 83.0 & 97.0 & 70.0& 88.0&92.7 \\
\bottomrule
\end{tabular}
}
\end{table}

\paragraph{Modularity.} Our approach separates reasoning capacity (projector) from domain adaptation (LoRA). This enables a plug-and-play setup: different projectors can be trained for different domains (math, code, science) while sharing the same frozen backbone. At inference, users can select the appropriate projector for their task or disable it entirely for general-purpose use.

\paragraph{Exploration via stochasticity.} The Gumbel-Softmax mechanism introduces controlled stochasticity during both training and inference. Our results show this benefits pass@$k$ at higher $k$, as the model explores diverse reasoning paths across samples. This contrasts with full vocabulary soft thinking, which achieves higher pass@1 but lower pass@16 due to more deterministic selections.

\section{Conclusion}

We introduced Soft Latent Thinking, a method that enables language models to reason in continuous embedding space through a lightweight projector operating over a compressed vocabulary. By replacing the full vocabulary projection with a small learned projector, we reduce per-step FLOPs during chain-of-thought while achieving competitive or superior performance.

Our experiments on DeepSeek-R1-Distill-Qwen-1.5B and LLaMA-3.2-3B-Instruct demonstrate that Soft Latent Thinking outperforms SofT-GRPO and other baselines at higher pass@$k$, with improved token efficiency. The approach requires only LoRA adapters on a frozen backbone, preserving out-of-domain capabilities and enabling modular deployment.

Our analysis reveals that the projector learns subtle but impactful adjustments to token selection, and that the compressed vocabulary introduces beneficial stochasticity for exploring diverse reasoning paths.

\section{Limitations}
The projector is initialized on domain-specific tokens (e.g., mathematical vocabulary), which limits out-of-domain transfer when the projector is enabled. On HumanEval (code), performance degrades with the math-initialized projector, though disabling it and using standard soft thinking with LoRA preserves performance. Future work could explore multi-domain projectors and better domain-agnostic initializations. A preliminary PCA/SVD initialization over the embedding table performed poorly in our early experiments, suggesting that useful reasoning traces may occupy a specialized subspace rather than a high-variance global embedding subspace.

\bibliography{references}

\end{document}